\documentclass[conference]{IEEEtran}
\makeatletter
\def\ps@headings{%
\def\@oddhead{\mbox{}\scriptsize\rightmark \hfil \thepage}%
\def\@evenhead{\scriptsize\thepage \hfil \leftmark\mbox{}}%
\def\@oddfoot{}%
\def\@evenfoot{}}
\makeatother
\usepackage[utf8]{inputenc}
\usepackage{multirow}
\usepackage[inline]{enumitem}
\usepackage{xcolor}
\usepackage{booktabs}
\usepackage{hyperref}
\usepackage{amsmath}
\usepackage{amssymb}
\usepackage{graphicx}
\usepackage[numbers]{natbib}
\usepackage{subfig}
\usepackage{tikz}

\usepackage[nomain, toc, acronym]{glossaries}
\glsdisablehyper

\usepackage{mathtools}
\newcommand\givenbase[1][]{\:#1\lvert\:}
\let\given\givenbase

\DeclarePairedDelimiterX\Basics[1](){\let\given\sgiven #1}

\usepackage{tabularx}

\newcommand\clearrow{\global\let\rowmac\relax}
\clearrow

\begin{document}

\title{SparseIDS: Learning Packet Sampling with Reinforcement Learning}

\author{\IEEEauthorblockN{Maximilian Bachl, Fares Meghdouri, Joachim Fabini, Tanja Zseby}
\IEEEauthorblockA{Technische Universität Wien\\
firstname.lastname@tuwien.ac.at}}




\newcommand\copyrighttext{%
  \footnotesize \textcopyright 2020 IEEE. Personal use of this material is permitted.
  Permission from IEEE must be obtained for all other uses, in any current or future
  media, including reprinting/republishing this material for advertising or promotional
  purposes, creating new collective works, for resale or redistribution to servers or
  lists, or reuse of any copyrighted component of this work in other works.}
\newcommand\copyrightnotice{%
\begin{tikzpicture}[remember picture,overlay]
\node[anchor=south,yshift=10pt] at (current page.south) {\fbox{\parbox{\dimexpr\textwidth-\fboxsep-\fboxrule\relax}{\copyrighttext}}};
\end{tikzpicture}%
}

\maketitle%
\copyrightnotice

\newacronym{ml}{ML}{Machine Learning}
\newacronym{dl}{DL}{Deep Learning}
\newacronym{aml}{AML}{Adversarial Machine Learning}
\newacronym{ids}{IDS}{Intrusion Detection System}
\newacronym{rnn}{RNN}{Recurrent Neural Network}
\newacronym{fgsm}{FGSM}{Fast Gradient Sign Method}
\newacronym{cw}{CW}{Carlini-Wagner method}
\newacronym{pgd}{PGD}{Projected Gradient Descent}
\newacronym{pdp}{PDP}{Partial Dependence Plot}
\newacronym{ars}{ARS}{Adversarial Robustness Score}
\newacronym{ttl}{TTL}{Time-to-Live}
\newacronym{dos}{DoS}{Denial-of-Service}
\newacronym{iat}{IAT}{Interarrival time}
\newacronym{rl}{RL}{Reinforcement Learning}
\newacronym{dpi}{DPI}{Deep Packet Inspection}
\newacronym{ac}{AC}{Actor-Critic}

\newcommand{\ours}{SparseIDS}

\begin{abstract}

\glspl{rnn} have been shown to be valuable for constructing \glspl{ids} for network data. They allow determining if a flow is malicious or not already before it is over, making it possible to take action immediately. However, considering the large number of packets that has to be inspected, for example in cloud/fog and edge computing, the question of computational efficiency arises. We show that by using a novel \gls{rl}-based approach called \textit{\ours{}}, we can reduce the number of consumed packets by more than three fourths while keeping classification accuracy high. To minimize the computational expenses of the \gls{rl}-based sampling we show that a shared neural network can be used for both the classifier and the \gls{rl} logic. Thus, no additional resources are consumed by the sampling in deployment. Comparing to various other sampling techniques, \ours{} consistently achieves higher classification accuracy by learning to sample only relevant packets. A major novelty of our \gls{rl}-based approach is that it can not only skip up to a predefined maximum number of samples like other approaches proposed in the domain of Natural Language Processing but can even skip arbitrarily many packets in one step. This enables saving even more computational resources for long sequences. Inspecting \ours{}'s behavior of choosing packets shows that it adopts different sampling strategies for different attack types and network flows. Finally we build an automatic steering mechanism that can guide \ours{} in deployment to achieve a desired level of sparsity.

\end{abstract}


\maketitle

\section{Introduction}
\label{sec:introduction}


With constant growth of today's cloud, fog and edge computing infrastructure the need for reliable and explainable anomaly and intrusion detection methods goes hand in hand. This is especially important for cloud, fog and edge computing since hosting companies commonly run third parties' code and hosting illegal activity can come with legal repercussions and is ethically discouraged. 

In the past, a viable way to detect intrusions was to analyze the contents of the packets themselves and determine, for example, whether a packet contains potentially harmful content by matching patterns. More recently, with the increasing deployment of encryption, the focus is now on features available for network monitoring devices even if packets are encrypted, such as packet sizes and temporal behavior. In our work we concentrate on features available when encrypting above transport layer (like for example TLS or QUIC), i.e. we also include port numbers and protocol flags.


Network communication is typically aggregated into \textit{flows}, which are commonly defined as a sequence of packets that share certain properties. When analyzing flows, not only the aforementioned features are available but also features representing the temporal behavior of the individual packets. Various approaches have been proposed to extract flow features with which anomaly detection can be performed: Authors in \cite{meghdouri_analysis_2018} compare such approaches and show their efficacy.
While these approaches often work well, a major drawback is that the whole flow must be received first and only then anomaly detection can be applied, to reveal malicious flows. Thus, we design a network \gls{ids} that operates on a per-packet basis and determines whether a packet is anomalous based on the aforementioned features.
In this way, an \gls{rnn}-based \gls{ids} has the benefit of avoiding feature engineering procedures and let the classifier build high level features by itself.
\cite{hartl_explainability_2020} show that such an architecture has similar performance to traditional flow-based anomaly detection systems, but can still detect anomalies \textit{before} the flow ends. Furthermore they analyzed its robustness with respect to adversaries extensively.


Nevertheless, not only the accuracy and reliability of classification, but also the computational efficiency, are critical for practical use. This follows from the fact that a rationally-acting organization will only implement an \gls{ids} if the cost saved by preventing security breaches is higher than the cost of the \gls{ids} itself. The cost of an \gls{ids} can be divided into (a) the cost of purchase/installation, (b) the cost of operation and (c) the cost of maintenance. The goal we pursue here is to develop optimal sampling strategies that do not significantly degrade classification performance while being able to only process a small fraction of the original data. Specifically, a good sampling technique should:

\begin{enumerate}
\item \textbf{choose optimally}, taking only samples that contain the most information and skip the less relevant ones.
\item have a parameter that allows to \textbf{trade off} classification performance for sparsity (choosing fewer packets).
\item be \textbf{independent of the classifier} so that any classifier can be used and that the classifier doesn't have to be aware of the sampling strategy.
\item be \textbf{retrainable} so that the sampling can be continuously adapted depending on the current threat landscape.
\item be able to \textbf{skip arbitrarily} many packets since network flows can be very long but only the first couple of packets are needed to decide whether an attack occurred or not. 
\item be \textbf{lightweight}, so that computation efficiency is not lower because of the sampling overhead. 
\end{enumerate}


For this purpose we develop the \gls{rl}-based \gls{ids} \textit{\ours{}} that fulfills the above properties. We show that a significant number of packets can be skipped while the accuracy does not drop
considerably. Compared to other common sampling techniques, \ours{} performs better when being trained on the same number of flows.


The accuracy-sparsity tradeoff of \ours{} can be regulated by a reinforcement learning parameter that directly affects both quantities. However, as in some deployments, a certain sparsity has to be minimally achieved to limit the computational expense, we also develop a steering system that adapts the tradeoff parameter in a closed control loop. In this way the given budget of computation power is not exceeded.

To make \ours{} as lightweight as possible, our implementation allows to let the sampling and the classifier be implemented in the same neural network, meaning that no overhead is introduced. With this option disabled, the computational power required is $2\times$ of what would be required without the sampling. Thus, for this option, less than $50\%$ of packets must be sampled to gain a computational efficiency advantage. Fortunately, we can show that with shared weights the same accuracy can be achieved while having no computational overhead.  

To encourage reproducibility and facilitate experimentation, we publicly release the source code, the trained ML models, the data and the figures of this work\footnote{\url{https://github.com/CN-TU/adversarial-recurrent-ids/tree/rl}}.

\section{Related Work}


Sampling of network packets to minimize computational effort has been extensively studied. 
In \cite{estan_building_2004}, authors have developed a scheme that varies the sampling rate depending on the available resources, making the sampling framework more effective than conventional fixed-rate sampling. In view of the numerous sampling techniques that exist, the authors in \cite{zseby_sampling_2005} published an overview of common methods for packet sampling in IP networks.


As far as sampling for \glspl{ids} is concerned, several approaches have been proposed to sample packets retaining high detection performance. Authors in \cite{estan_new_2003} propose to single out large flows for \gls{ids} and sample fewer packets for increased computational efficiency. In \cite{bakhoum_intrusion_2011} an approach based on Markov chains to sample packets for \gls{ids} was used. \cite{murali_kodialam_detecting_2003} lay out a game-theoretic model for determining which network paths are more vulnerable and require more packet sampling and \cite{ha_suspicious_2016} develop a system for Software Defined Networks, which aims to sample more packets from more vulnerable parts of the network.


Considering recurrent approaches for network traffic, \gls{rnn}-based \glspl{ids} have been shown to perform very well. In \cite{hartl_explainability_2020}, authors build an LSTM-based recurrent classifier for Network Intrusion Detection and further develop explainability methods as well as methods for security assessment of recurrent classifiers.


Regarding skipping parts of sequences for \glspl{rnn}, some works investigated ways to let \glspl{rnn} skip parts of sequences. The focus of these works was mostly Natural Language Processing and we know of no work that focuses on network traffic. The following papers provide an overview of this research domain:

Authors in \cite{yu_learning_2017} propose the usage of \gls{rl} to make an LSTM network learn to skim text. Their technique does not aim to maximize sparsity but only lets the \gls{rl} optimize classification performance for the sequence as a whole. Thus, if it learns to skip elements it is only because it is better for achieving good classification performance, not because it wants to increase sparsity.

Another technique developed in \cite{campos_skip_2018} includes a \textit{skip gate} into LSTM and other recurrent cells. It works without \gls{rl} but on the downside, their method can only be trained once and not be adapted afterwards and also it explicitly depends on the the implementation of the underlying classifier, while we want a solution in which the sampling procedure and the classifier are independent.

An LSTM network with two state vectors (one full state vector and one reduced one) was used in \cite{seo_neural_2018}. At each step the network decides whether to use the reduced or the full state vector and can reduce computational cost by using the reduced one. For network traffic it is often not even necessary to have a reduced state vector since for some attack flows, already the first couple of packets might contain enough information and then the remaining packets of the flow do not have to be considered anymore at all \cite{hartl_explainability_2020}.

\cite{gui_long_2019} use \gls{rl} but their intention is not to skip samples but instead to make the \gls{rl} learn to choose a previous state that can aid at the current step. Thus it tries to help the \gls{rnn} with memorizing information from the past but does not actually take less data, which is the goal we pursue in this work.

These proposed techniques focus their evaluation on text and hence do not fulfill all criteria we set out in \autoref{sec:introduction}. Furthermore, all techniques that use \gls{rl} use discrete actions, meaning that there is a hyperparameter $k$, which influences the maximum jump that is possible. However, for a network flow, already after the second packet it might be obvious that a flow is an attack and thus the remaining packets can be completely ignored. It is therefore beneficial not to have a fixed maximum step size but a continuous one. This allows skipping arbitrarily many packets and gives more flexibility  to the sampling framework.


\section{\ours{}}

To fulfill the goals outlined in \autoref{sec:introduction}, in this section we propose the architecture of our proposed sampling framework \ours.

\subsection{Classifier}
\label{subsec:classifier}
We implemented a three-layer LSTM-based classifier with 128 neurons at each layer. We choose 128 neurons as we don't notice a significant degradation of performance compared to the results of \cite{hartl_explainability_2020} (check Table II, flow accuracy) and as fewer neurons allowed for more agility during the training phase and allowed us to run more experiments. As the input features we use source port, destination port, protocol identifier, packet length, \gls{iat} to the previous packet in the flow, packet direction (i.e. forward or reverse path) and all TCP flags (0 if the flow is not TCP).
We omitted \gls{ttl} values, as they are likely to lead to unwanted prediction behaviour \cite{bachl_walling_2019}.  Among the used features, source port, destination port and protocol identifier are constant over the whole flow while the others vary.
Additionally, we add the number of skipped packets since the last packet as a feature, which can help the classifier because otherwise it might be confused because of missing packets.
We used the usual 5-tuple flow key, which distinguishes flows based on the protocol they use and their source and destination port and IP address.

\subsection{Reinforcement Learning-based Sampling}
\label{subsec:rlForSampling}

Given enough training instances, \gls{rl} algorithms have been shown to perform very well in complex decision-making situations such as playing video games \cite{mnih_playing_2013} as well as networking problems such as congestion control \cite{bachl_rax_2019}. 
We thus consider an \gls{rl}-based architecture to be suitable for taking packet sampling decisions. The system should look at the current packet as well as the history of the conversation flow and output how many packets should be skipped until the next packet is considered in the current flow. The reward that the \gls{rl} tries to maximize should be a combination of classification accuracy and the number of packets that could be skipped (sparsity). An operator should be able to specify how much accuracy is allowed to be traded off for achieving higher sparsity.


For a flow of length $N$ at each packet $n$ we have two reward metrics which we compute in a fashion that is known from \textit{R-learning} \cite{schwartz_reinforcement_1993}:


\begin{figure*}

\centering
  \includegraphics[width=0.8\textwidth]{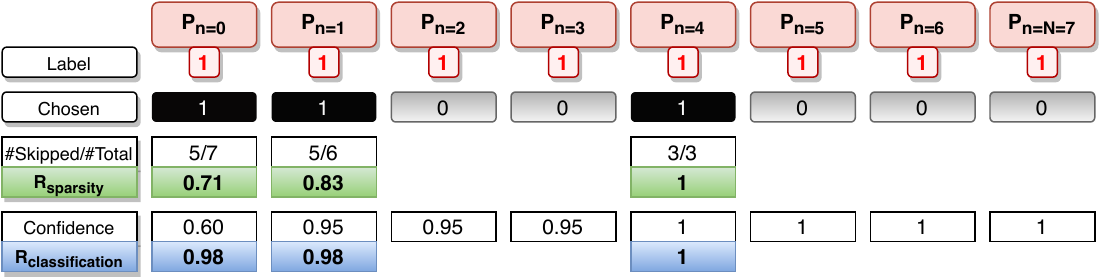}
  \caption{The computation of the rewards for a sample malicious flow: Gray packets (0) are skipped while black ones (1) are chosen. ``\#Skipped\//\#Total'' (number of packets skipped with respect to the total number of packets both starting from the current position) and ``Confidence'' are the underlying parameters for rewards calculations (see \autoref{eq:R_classification} and \autoref{eq:R_sparsity}). Numbers are rounded to two digits after the decimal point.}
  \label{fig:rewards}
\end{figure*}

The classification reward metric (\autoref{eq:R_classification}) captures how correct the classifier is on average for all future packets of a flow. The label of each packet in a flow has a value of 0 for non-attack traffic and 1 for attack traffic with one flow being only attack or non-attack but not mixed. The confidence is the sigmoided $\left(\frac{e^x}{1+e^x}\right)$ output of the classifier, yielding a value between 0 and 1, where 0 is absolute confidence for a flow being benign and 1 is absolute confidence for a flow being an attack.
\begin{align}
\begin{split}
R_{\text{classification},n} = \frac{1}{N-n} \sum_{i=n+1}^{N} 1 - |\text{label}_i - \text{confidence}_i|
\end{split}
\label{eq:R_classification}
\end{align}
Intuitively, the classification reward makes sure that packets are chosen that contain the most information and thus result in the highest classification performance.

The sparsity reward metric (\autoref{eq:R_sparsity}) captures how much sparsity is achieved on average for all future packets of a flow. E.g. for a flow of length 10 if after 5 packets have been processed, 3 packets are chosen and 2 are skipped from the remaining 5 packets, the sparsity reward metric at packet 5 is $\frac{2}{10-(4+1)}=\frac{2}{5}$.
\begin{align}
\begin{split}
R_{\text{sparsity},n} = \frac{\sum_{i=n+1}^{N} 1_{\text{skipped},i}}{N-(n+1)}
\end{split}
\label{eq:R_sparsity}
\end{align}
\autoref{fig:rewards} shows how classification and sparsity rewards are computed for a toy example.

\begin{figure}[h]
\centering
  \includegraphics[width=0.8\columnwidth]{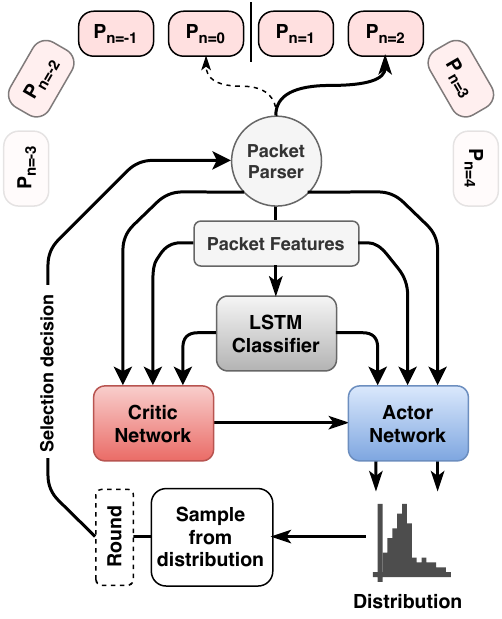}
  \caption{SparseIDS framework. Three deep networks are shown: the classifier, the actor network and the critic network. The actor takes actions based on the feedback from the the critic network and the classifier. The critic network tracks the actor's actions based on the classifier confidence. The packet selector chooses packets based on the decision of the actor network.}
  \label{fig:sampling}
\end{figure}

We opt for an \emph{\gls{ac}} approach (as shown in \autoref{fig:sampling}) modeled after \cite{mnih_asynchronous_2016} because we consider it being more interpretable than a Deep \gls{rl} approach based on a \textit{Deep Q Network} like \cite{mnih_playing_2013}, as it outputs the reward that was expected and the reward that was actually achieved, which we explain in more detail in the next paragraphs.

\begin{figure}[h]
\centering
  \includegraphics[width=\columnwidth]{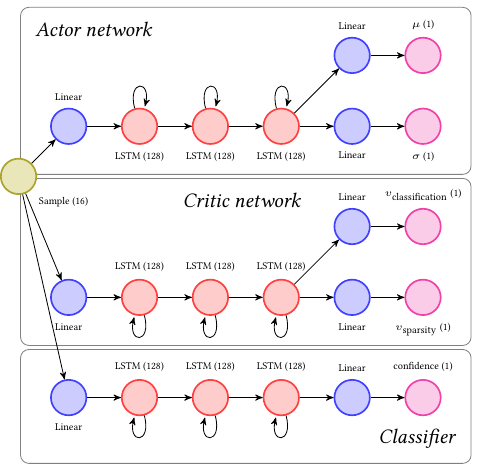}
  \caption{An overview of the complete neural network being used in our experiments. Ocher stands for the input, blue for linear layers, red for LSTM layers and purple for the outputs of the neural network. The numbers in parentheses stand for the width of that layer. Our implementation also supports sharing the LSTM between all three neural networks, which greatly reduces computational complexity: In this case, the first linear layers as well as the three layers of LSTM cells are shared while the last linear layers are still separate. We later show that shared weights perform on par with separate ones.}
  \label{fig:neuralNetworkArchitecture}
\end{figure}

Our proposed framework (see \autoref{fig:neuralNetworkArchitecture}) consists of three independent neural networks: the classifier, the critic and the actor.

The classifier's goal is to guess correctly if a flow is an attack or not. For this, it outputs its confidence for non-attack/attack after reading a packet. As stated in \autoref{subsec:classifier}, we choose an LSTM-based supervised classifier for this work, but our framework can work together with any classifier as long as it outputs a confidence for non-attack/attack at each packet and can handle skipped packets. For instance, it would also be possible to replace the LSTM with an unsupervised autoencoder akin to \cite{mirsky_kitsune_2018}.

The critic aims to estimate the future expected average classification performance and the future expected average sparsity that can be achieved at a packet $n$. For this, given the current input vector and the state of the LSTM cells $s_n$ and the neural network weights of the critic $\theta_v$, it outputs the value functions $v_{\text{classification},n}$ and $v_{\text{sparsity},n}$ at each packet $n$. It aims to predict the classification performance and sparsity correctly for each packet by minimizing the following loss function:
\begin{align}
\begin{split}
l_{\text{v},n} =& \left(R_{\text{classification},n} - v_{\text{classification},n}\left(s_n ; \theta_\text{c} \right)\right)^2 \\
+ & \left(R_{\text{sparsity},n} - v_{\text{sparsity},n}\left(s_n ; \theta_\text{c} \right)\right)^2
\end{split}
\end{align}

Now that the reward metrics and the critic's outputs are defined, we can define the overall utility $U$ that we want to maximize at each packet $n$ using the sparsity-accuracy tradeoff parameter $\alpha$ as follows:
\begin{align}
\begin{split}
U_n =& \left(R_{\text{classification},n} + \alpha \cdot R_{\text{sparsity},n}\right) \\
- & \left(v_{\text{classification},n} + \alpha \cdot v_{\text{sparsity},n}\right)
\end{split}
\end{align}

This means that we want the reward to be higher than the critic's expectation. Note that it would be also possible to unify both reward metrics in one and have the critic have only one output, but we consider it to be more interpretable to have two separate outputs, which can be inspected by an operator.

The actor outputs a probability distribution at packet $n$, which is sampled to determine the number of packets that should be skipped next. It aims to change the distribution so that actions which result in a higher reward are chosen more frequently, maximizing the utility function. However, this strategy could lead to the actor getting stuck with what it considers the best decision and never try alternatives even though they might result in a higher reward. This would be equivalent to the actor being stuck in a local maximum. Thus, the actor not only tries to change the distribution so that the actions become more optimal (higher reward) but at the same time also aims to maximize the entropy of the distribution so that alternative choices are still explored reasonably often.

Specifically, at packet $n$ of a network flow, given the current state $s_n$ (which consists of the current input as well as the LSTM state) and the neural network weights of the actor $\theta_a$, the actor outputs a probability distribution $\pi \left( a_n \given s_n ; \theta_\text{a} \right)$ from which the action $a_n \ge 1$ is sampled. $a_n=1$ means that the next packet is inspected next, while, for example, $a_n=2$ means that one packet is skipped until the next that is chosen, $a_n = 3$ means that two are skipped and so forth. Besides wanting to make actions which result in a high reward more likely, the actor also seeks to optimize the entropy $H(\cdot)$ of the probability distribution so that it keeps exploring and doesn't get stuck with a suboptimal policy.

Combining these objectives, the actor network aims to minimize the loss function
\begin{align}
\begin{split}
l_{\text{a},n} =& \underbrace{ -\log \left( \pi \left( a_n \given s_n ; \theta_\text{a} \right) \right) U_n}_{\text{Policy Loss}} - \beta \underbrace{H\left( \pi\left( s_n; \theta_a \right)\right) }_{\text{Entropy}}
\end{split}\label{eq:actor}
\end{align}


In related work regarding \gls{rl}-based sampling for sequences, only discrete actions were evaluated so far: The actor would output a categorical probability distribution with a fixed maximum number of bins $k$. A drawback of this approach is that only a predefined maximum number of packets can be skipped at once (the maximum of the actions space) and that for a very large number of $k$, it can take a long time to converge to an optimum policy as there are $k$ different options to explore at each packet $n$. Besides discrete actions, we thus also experiment with continuous actions, which allow to skip arbitrarily many packets. 

\subsection{Continuous Actions}

For performing continuous actions we need a probability distribution which can be parametrized to have the mean between 0 and positive infinity. Furthermore it must have at least two parameters so that the standard deviation, which is required for the entropy (\autoref{eq:actor}), can be changed independently of the mean. For example, the Exponential distribution can be parametrized to have its mean anywhere between 0 and positive infinity, however, it has only one parameter and thus the mean and the standard deviation cannot be independently changed. Conversely, a simple distribution which fits our requirements is the log-normal distribution, which is defined as a normal distribution whose output is exponentiated.


The architecture in \autoref{fig:neuralNetworkArchitecture} shows that the actor network has two outputs: $\mu$ and $\sigma$. These encode the mean and standard deviation of a log-normal distribution and can never be smaller or equal to zero and hence we always apply a rectifier function at the output. We choose the softplus ($\log\left(1 + e^x \right)$), which is a smooth rectifier function:
When sampling from the log-normal distribution, a real number is obtained. We round it down and then add 1. This gives us a natural number between 1 and $\infty$, which states how many packets should be skipped. 

When experimenting with continuous actions, we encountered one practical problem: At the beginning of training, the classifier would still perform badly and would not give the actor a useful reward. As a result, the actor would only focus on sparsity and aim to maximize it, as the classification reward would always be equally bad no matter how many packets are chosen. Thus, the actor would learn to let $\mu$ go to infinity. This would increase the reward at the beginning when the interaction between the three components of our \gls{rl} system is still fragile. As a solution, we add a penalty to the reward calculation when the actor jumps far behind the last flow packet. The modified reward applies only to the last chosen packet and penalizes the actor jumping more than one packet behind the last packet. The sparsity reward formula (\autoref{eq:R_sparsity}) of the last packet is modified as follows, where \textit{last} refers to the index of the last chosen packet in the flow ($n=\textit{last}$):
\begin{align}
R_{\text{sparsity},\textit{last}} = \frac{\left(\sum_{i=\textit{last}+1}^{N} 1_{\text{skipped},i}\right)}{N- (\textit{last}+1+\left(\textit{last}+a_{\textit{last}} - N - 1\right))}
\label{eq:R_sparsity_last}
\end{align}
Specifically, it adds the term $\left(\textit{last}+a_{\textit{last}} - N - 1\right)$ to the denominator which penalizes the actor jumping more than one packet behind the flow. Jumping just right after the flow does not get penalized because if the actor considers the rest of the flow as irrelevant, ignoring the rest of the flow and jumping to the first packet behind the flow is a useful action.

\section{Other sampling techniques}
\label{sec:other_sampling_techniques}
In order to evaluate the performance of \ours, we compare it with a few straightforward sampling techniques.
We opt for three sampling families that, unlike \ours{} cannot adapt their strategy based on packet contents. 

\subsection{Random Sampling}

Given the fraction $p$ of packets to be selected in a sequence of $N$ packets, this technique will sample each individual packet with probability of $p$. Theoretically, if the sequence is large enough (infinitely large), $m$ packets will be randomly sampled from a uniform distribution whereby $m \approx N \times p$. Naturally, the number of chosen packets can vary to some degree. This variation is larger for short flows. In practice, random sampling works well if several packets have similar properties (redundancy) and therefore the probability of skipping important information is minimal.

\subsection{Relative First \emph{m} Packets}
As its name suggests, this technique only samples the first packets from each flow and ignores the rest of the flow. For a sampling rate $p$ it will take the first $m$ packets from each flow so that $m \approx N \times p$. For example, for a flow with length 10 and $p=0.2$, it will take the first 2 packets, for another flow of length 40 and for the same sampling rate ($p=0.2$), it will sample the first 8 packets. For this procedure the length of a flow has to be known beforehand and thus it is not applicable in a streaming scenario. This technique is successful under the assumption that most information for determining maliciousness is present in the first few packets of a flow, i.e. a large data transfer in which the initial packets represent a hand-shake after which the majority of packets contain the raw data and do not reveal significant information to non-\gls{dpi} systems.

\subsection{First \emph{m} Packets}
This technique is similar to the one above, with the difference that the length of a flow doesn't have to be known beforehand. If flows have an average length of 10 packets, and the sampling rate $p=0.2$ it would take up to two packets from each flow regardless of its length. To determine the average length of flows, it is necessary to use experience from past flows. 

\subsection{Every \emph{i}th Sampling}
Finally, we use a periodic sampling technique that, for a sampling rate $p$, takes a sample each $i \approx \frac{1}{p} $ packets. Similar to the last technique, the length of the flow is not needed in this case. In practice, this technique is best suited to keep the distribution of the sampled set similar to the original distribution, and it is thus suited for scenarios where information is heterogeneously distributed between packets over an entire network flow.

In each of the above techniques, the first packet of each flow is always taken because otherwise one could get undefined behavior: If a flow only consists of 1 packet and this packet is not chosen, it is not even defined if the prediction of the classifier was correct for this flow or not. Thus, in our implementations of the above sampling techniques, we make sure that even when the first packet is always chosen, the overall fraction of chosen packets corresponds to the sampling probability $p$.

After applying the sampling techniques, the sequence of selected packets is then fed to the classifier discussed in \autoref{subsec:classifier}, where features are extracted and a decision is made after each packet. Also, the classifier gets information on how many packets were skipped prior to each packet.


\section{Dataset}

For our experiments, we use the \textit{CIC-IDS-2017} \cite{sharafaldin_toward_2018} dataset, which includes more than 2 million flows of network data, containing both benign (74.75\%) traffic and a large number of different attacks (25.25\%, 14 attack types). 

We use Z-score normalization and a train/test split of 2:1. To speed up experimentation, we cut flows at a maximum length of 20. As Figure \ref{fig:inspectingAll} shows, there are only few flows that exceed that length.

\section{Results}

\begin{figure}[h]
\centering
  \includegraphics[width=\columnwidth]{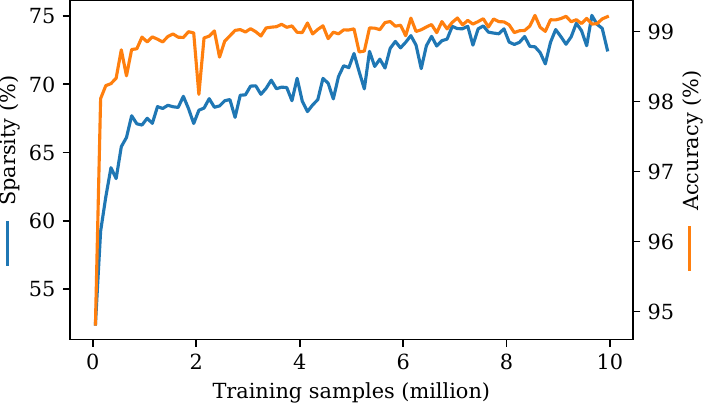}
  \caption{Training with a tradeoff of 0.1 with continuous actions for several epochs. Training accuracy increases sharply in the beginning while increasing more slowly later on. The behavior for sparsity is similar. As the classifier gets better, the \gls{rl} starts skipping more packets.}
  \label{fig:training}
\end{figure}


For all our experiments we train our neural networks for 8 epochs ($>10$ million flows) with the Adam optimizer with a learning rate of $0.001$. An important implementation detail is that we always start with the first packet of a flow, meaning that it cannot be skipped. Also, we set the entropy parameter to $0.01$. 
\autoref{fig:training} shows how, during training, accuracy increases quickly at the beginning and then gradually afterwards. At the same time, sparsity gets higher as the \gls{rl} learns which packets can be skipped since they do not influence classification performance. 

Opposed to the behavior during training, we do not randomly sample from the probability distribution of the actor during deployment. We instead take the mean of the distribution for the continuous case and the mode for the discrete one. This is reasonable and best practice \cite{mnih_asynchronous_2016} since during deployment we do not aim to explore the possible choices anymore as they were already learned during training: Only the optimal choice, which was learned, should be applied and no exploration is necessary since the actor does not train anymore. Furthermore, the critic is only necessary during training for the reward of the actor. During deployment, it can be disabled, saving computational power.

\subsection{Influence of the tradeoff}
\begin{table}[h]
\caption{Comparing classification metrics per flow with different tradeoffs. The higher the tradeoff, the higher the sparsity but also the lower the accuracy.} \label{tab:results_tradeoff}
\centering
\begin{tabular*}{\columnwidth}{>{\rowmac}l @{\extracolsep{\fill}} >{\rowmac}c>{\rowmac}c>{\rowmac}c>{\rowmac}c<{\clearrow}} \toprule
Tradeoff & 0.0 & 0.1 & 0.5 & 1.0 \\	\midrule
Sparsity & 0\% & 76.3\% & 76.5\% & \textbf{76.8\%} \\ \midrule
Accuracy & \textbf{99.5\%} & 99.4\% & 99.3\% & 99.0\% \\
Precision & \textbf{99.4\%} & 98.5\% & 98.7\% & 97.9\% \\
Recall & 98.5\% & \textbf{99.0\%} & 98.6\% & 98.1\% \\
F1 & \textbf{99.0\%} & 98.7\% & 98.6\% & 98.0\% \\
Youden & 98.4\% & \textbf{98.5\%} & 98.2\% & 97.4\% \\
\bottomrule
\end{tabular*}
\end{table}
\autoref{tab:results_tradeoff} shows the influence of the tradeoff parameter $\alpha$. For columns with a tradeoff of 0.0, we train the classifier regularly with all packets without using \gls{rl}. For tradeoffs 0.1, 0.5 and 1.0 we train \ours{} using \gls{rl} with continuous actions and keep the tradeoff constant for the whole training. Results show that the higher the tradeoff, the fewer packets are sampled and the lower the accuracy.

\subsection{Discrete vs.~Continuous actions}

\begin{table}[h]
\caption{Comparing classification metrics per flow with either discrete (20 different actions) or continuous actions. The tradeoff is always $0.1$.} \label{tab:results_disc_cont}
\centering
\begin{tabular}{>{\rowmac}l >{\rowmac}r>{\rowmac}r<{\clearrow}} \toprule
Actions & continuous & discrete \\	\midrule
Sparsity & \textbf{76.3\%} & 73.6\% \\ \midrule
Accuracy & \textbf{99.4\%} & 99.2\% \\
Precision & 98.5\% & \textbf{98.7\%} \\
Recall & \textbf{99.0\%} & 98.3\% \\
F1 & \textbf{99.7\%} & 98.5\% \\
Youden & \textbf{98.5\%} & 97.8\% \\
\bottomrule
\end{tabular}
\end{table}
\autoref{tab:results_disc_cont} compares continuous and discrete (20 different) actions and shows that continuous actions result in a higher accuracy with fewer samples being selected. We chose 20 discrete actions as flows have a maximum length of 20, so with 20 actions, the \gls{rl} can decide to skip the entire flow at the first packet. When we tried with fewer discrete actions (like 8), we saw that sparsity was limited as the \gls{rl} would have liked to skip more packets but couldn't. The results lead to the conclusion that at least for our use case and for the dataset we consider, continuous actions seem to perform better. Generally, the advantage of discrete actions is that they are theoretically more powerful as they can compare each of the possible actions with each other, while the only possibility for continuous actions is to increase or decrease the mean of the log-normal distribution, which can lead to worse performance. On the other hand, with discrete actions, the maximum number of samples to skip is fixed, which we consider to be a major drawback. Furthermore, training time is potentially higher since, for example, for a discrete action space of 20 actions, each option is treated as being completely different and, while actually there is not a large difference between skipping 18 or 19 packets, there is a big difference between skipping 0 packets compared to skipping 19 packets. This means that for discrete actions, all actions are perceived as being completely independent, which is not true as the reward function considers sparsity, which increases gradually as more packets are skipped and thus similar actions result in similar rewards.

\subsection{Comparing with other sampling methods}

\begin{table}[h]
\caption{Comparing classification metrics per flow between \gls{rl}-based sampling with continuous actions (tradeoff 0.1) and alternative strategies.}
\label{tab:results_others}
\centering
\begin{tabular*}{\columnwidth}{>{\rowmac}l @{\extracolsep{\fill}} >{\rowmac}c>{\rowmac}c>{\rowmac}c>{\rowmac}c>{\rowmac}c<{\clearrow}} \toprule
& \gls{rl} & random & every $i$th & relative first \emph{m} & first \emph{m} \\	\midrule
Sparsity & 76.3\% & 76.3\% & 76.3\% & 76.3\% & 76.3\% \\ \midrule
Accuracy & \textbf{99.4\%} & 96.9\% & 97.8\% & 97.3\% & 98.3\% \\
Precision & \textbf{98.5\%} & 92.4\% & 95.8\% & 93.8\% & 95.6\% \\
Recall & \textbf{99.0\%} & 95.6\% & 95.4\% & 95.5\% & 97.9\% \\
F1 & \textbf{99.7\%} & 94.0\% & 95.6\% & 94.7\% & 96.7\% \\
Youden & \textbf{98.5\%} & 93.0\% & 93.9\% & 93.4\% & 96.4\% \\
\bottomrule
\end{tabular*}
\end{table}

Next we compared the \gls{rl} technique used by \ours{} with several common sampling strategies suggested in \autoref{sec:other_sampling_techniques}.
\autoref{tab:results_others} reveals that \ours{} achieves the highest accuracy with the same sparsity. Not surprisingly, random sampling performs the worst, as it is likely that a flow is ``unfortunate'' and doesn't get enough packets sampled while another one gets more packets or the packets that carry more useful information. The best strategy besides the \gls{rl} approach seems to be ``first $m$'' sampling, which matches the conclusion from \cite{hartl_explainability_2020} that the first couple of packets are the most important ones for determining maliciousness of a flow.
``first $m$'' performs better than ``relative first $m$'' because for ``relative first $m$'', longer flows will have more packets sampled from them. Conversely, very short flows will get few packets. For example, with $p=0.2$, a flow of length 20 will get 4 packets while one with 4 packets will get only 1. However, with ``first $m$'', the short and the long flow will get the same number of packets. Apparently it is better if all flows get the same number of packets than if long flows get more packets. 

\subsection{Shared weights}
\begin{table}[h]
\caption{Using shared weights for all components of \ours{} resulting in no overhead in deployment compared to a classifier without our \gls{rl} framework.} \label{tab:results_shared}
\centering
\begin{tabular*}{\columnwidth}{>{\rowmac}l @{\extracolsep{\fill}} >{\rowmac}c>{\rowmac}c>{\rowmac}c>{\rowmac}c<{\clearrow}} \toprule
Tradeoff & 0.0 & 0.1 & 0.5 & 1.0 \\	\midrule
Sparsity & 0\% & 77.6\% & 77.6\% & \textbf{78.2\%} \\ \midrule
Accuracy & \textbf{99.5\%} & 99.4\% & 99.3\% & 99.0\% \\
Precision & \textbf{99.4\%} & 99.2\% & 98.6\% & 98.9\% \\
Recall & \textbf{98.5\%} & 98.2\% & 98.7\% & 96.9\% \\
F1 & \textbf{99.0\%} & 98.7\% & 98.6\% & 97.9\% \\
Youden & \textbf{98.4\%} & 97.9\% & 98.2\% & 96.6\% \\
\bottomrule
\end{tabular*}
\end{table}
\autoref{tab:results_shared} shows that when sharing the neural network weights of the classifier, the actor and the critic, \ours{} achieves the same very high accuracy as in the case in which all every component has its own weights (\autoref{tab:results_tradeoff}). The achieved sparsity is even higher when using separate weights. One explanation for this might be that sharing weights has a regularizing effect. 

With shared weights, the \gls{rl} mechanism adds no computational complexity since all the operations in the neural network are the same for the actor and the classifier (the critic is not needed during deployment). The only part that is not shared is the last linear layer, which connects the last LSTM layer with the outputs.

\subsection{Inspecting the \gls{rl}'s behavior}
\label{subsec:inspecting}

\begin{figure*}[h]
\centering
\subfloat[Benign Traffic\label{fig:inspectingNormal}
]{
\includegraphics[width=0.98\columnwidth]{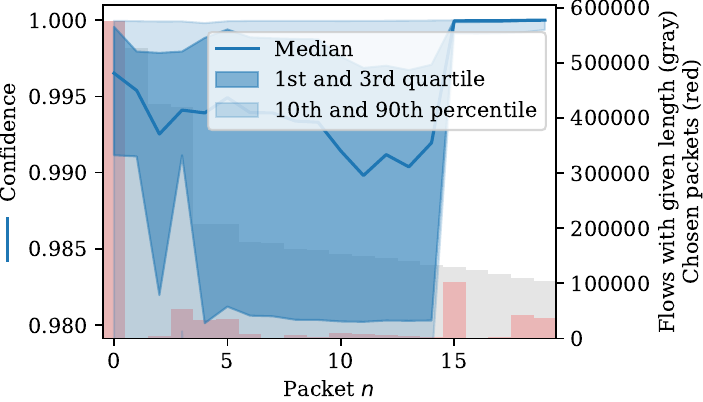}
}{}
\subfloat[SSH Patator\label{fig:inspectingPatator}
]{
\includegraphics[width=0.98\columnwidth]{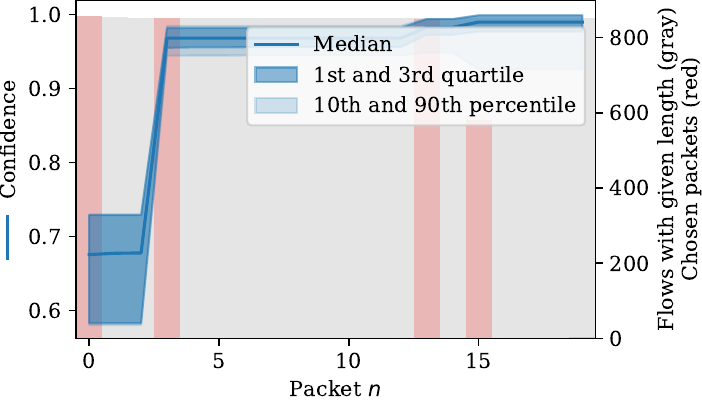}
}{}
\subfloat[DDoS Slowloris\label{fig:inspectingSlowloris}
]{
\includegraphics[width=0.98\columnwidth]{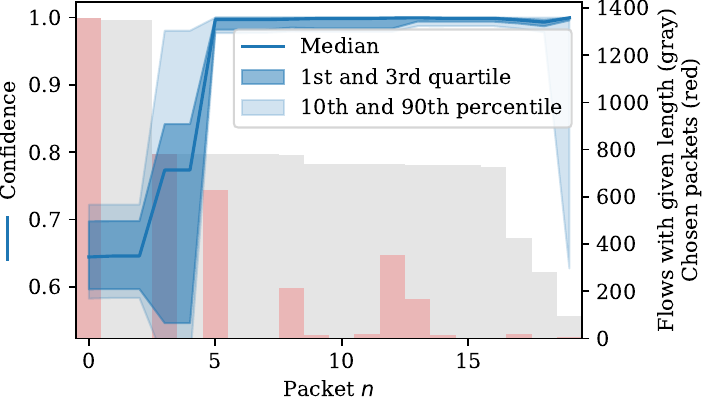}
}{}
\subfloat[All samples\label{fig:inspectingAll}
]{
\includegraphics[width=0.98\columnwidth]{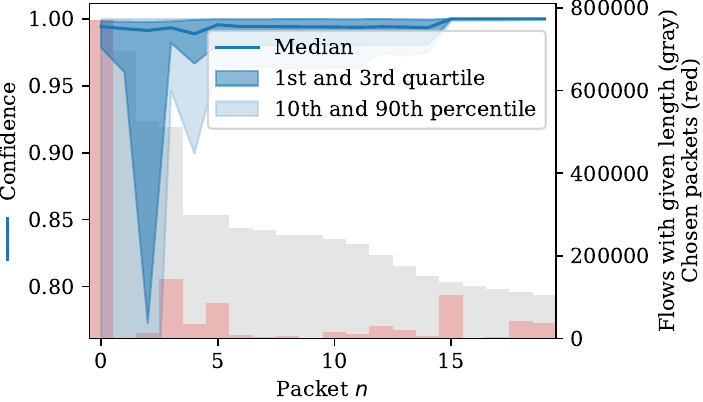}
}{}
\caption{Inspecting the sampling strategy of \ours{} for various attack types with continuous actions and a tradeoff of $0.1$. The blue line shows the confidence. The gray bars show how many samples had this number of packets. For example, the gray bar in Figure \ref{fig:inspectingNormal} at position 19 with height 10000 signifies that 10000 flows of benign traffic had a length of at least 20 packets (since we say that the first packet is at position 0). The red bars represent the number of packets sampled at this position for all flows. The first packet is always chosen for each flow by design.}
\label{fig:inspecting}
\end{figure*}


\autoref{fig:inspecting} shows the sampling strategy for different attack types. Figure \ref{fig:inspectingNormal} shows that for benign traffic, the sampling strategy is quite heterogeneous: Not specific packets get always chosen, but different packets get chosen, depending on the flow. This contrasts with SSH Patator (Figure \ref{fig:inspectingPatator}), where the 0th, the 3rd, the 13th and the 15th packet are always chosen, since they appear to be especially indicative of an attack. Also, the flows seem to contain no useful information between the 3rd and the 13th packet, so that this part is always skipped altogether. Slowloris (Figure \ref{fig:inspectingSlowloris}) is more similar to what we saw for benign traffic: The sampling strategy varies a lot for different flows and not always the same pattern is used. When looking at all flows (Figure \ref{fig:inspectingAll}) it becomes apparent that the second packet (index 1) is almost never chosen, and the third is also chosen quite rarely. We attribute this to the fact that most communication flows are TCP and use handshakes, meaning that the second packet is simply a reply a part of the handshake, revealing no important information. In general, we see that the sampling strategy varies significantly between different attack types and for some attack types even for different flows within the same attack type, showing that \ours{} learns to choose the packets effectively, which reveal the malicious nature of a flow.


\section{Steering}

So far, we have always trained \ours{} with a fixed tradeoff parameter $\alpha$. Nevertheless, there is also an important use case in which an operator wants a certain minimum sparsity, for example 50\%, to limit the necessary computation power and thus reduce costs.
We thus propose a method that enables steering the tradeoff to achieve a certain desired sparsity. The steering algorithm works as follows:
\begin{enumerate}
\item A model is trained for which the tradeoff is chosen randomly for each flow during training from a uniform distribution yielding values between 0 and 1. We add a new feature to each packet that indicates the tradeoff chosen for the respective flow so that the \gls{rl} can adjust its sampling strategy in anticipation of the tradeoff that will determine the reward.
\item During deployment, the tradeoff is set to the maximum that was used during training (1 in our case), which results in maximum sparsity. The tradeoff is then continuously decremented if the current sparsity is still above the specified minimum. It is important that the tradeoff can never fall below 0 because this is a case that has not been encountered during training and can lead to abnormal behavior.
\end{enumerate}

\begin{figure}[h]
\centering
  \includegraphics[width=\columnwidth]{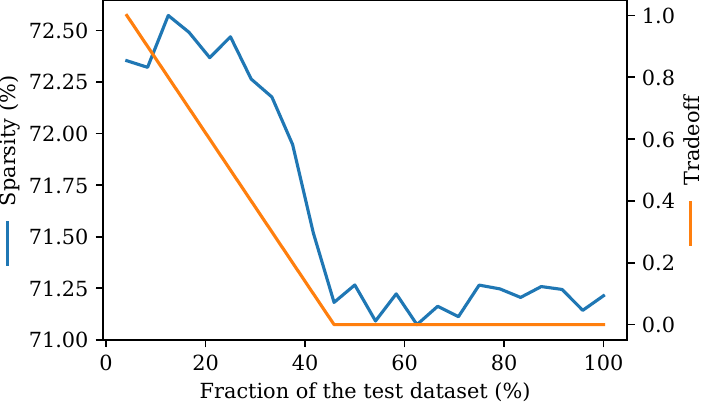}
  \caption{Steering the tradeoff with an allowed minimum sparsity of 50\% over the test dataset. The tradeoff is initialized as 1 and continuously lowered in a streaming fashion as long as the sparsity is higher than 50\%. The procedure stops if the minimum sparsity is reached or the tradeoff reached its minimum of 0, which is the case here.}
  \label{fig:steering}
\end{figure}


\autoref{fig:steering} shows the steering in action. We start with a maximum sparsity of approx.~72.5\% and try to reach 50\%. To do so, the steering system starts decreasing the tradeoff parameter until it reaches 0. It might seem surprising that even with a tradeoff of 0, which means that only classification matters and sparsity has zero influence, a sparsity of 73.5\% is still achieved. We attribute this to the fact that \ours{} learns that some flows may contain packets that do not contribute to determining maliciousness at all and are skipped since they do not affect the accuracy. 

\section{Conclusion}
\label{sec:conclusion}
In this paper we aimed to develop smart packet sampling methods for the use case of \glspl{ids} to save computational power. To this end, we proposed \ours{} which is a framework for packet sampling based on \gls{rl}. We showed that the \gls{rl}-based approach with a tradeoff between classification accuracy and sparsity indeed results in higher sparsity the larger the tradeoff parameter, while potentially classification performance is sacrificed. Surprisingly, we saw that \ours{} on average chooses very few packets per flow: Typically $\approx \frac{1}{4}$. Furthermore, it outperforms non-smart sampling techniques.

Besides discrete actions for our \gls{rl} system we also implemented continuous ones. We showed that a naive implementation would not work as the training is too fragile in the beginning and the \gls{rl} system learned to maximize only one of the rewards (sparsity) in the beginning. This was solved by adding a penalty for jumping behind the end of the sequence which proved effective. Comparing with discrete actions, continuous actions seem to perform slightly better given the same training time. Additionally it is also an advantage that for continuous actions the neural network only needs to have two outputs (mean and standard deviation) while for discrete ones it needs as many outputs as there are actions. The continuous action framework is thus more lightweight and also can learn to skip arbitrarily many packets.

We showed that using a shared neural network for both the \gls{ids} as well as the sampling can achieve the same classification accuracy and sparsity while not resulting in any computational overhead during deployment. 

Inspecting the decisions of our \gls{rl} framework showed that the sampling strategy is clearly different for different attack strategies and also within some attack families, indicating that \ours{} indeed adopts different sampling strategies for different conversation flows.

Eventually, we implement a steering method to help achieving a certain sparsity threshold. The steering mechanism can be used to limit the computational expenses of an operator.

In general, we think that \gls{rl}-based packet sampling is not only applicable to \glspl{ids} but we are confident that other domains of networking can benefit from similar solutions to save computational expenses.

\section*{Acknowledgements}
The Titan Xp used for this research was donated by the NVIDIA Corporation.

\renewcommand*{\bibfont}{\small}
\bibliographystyle{ieeetr}
\bibliography{bibliography}

\end{document}